\documentclass[conference]{IEEEtran}
\IEEEoverridecommandlockouts
\usepackage{cite}
\usepackage{amsmath,amssymb,amsfonts}
\usepackage{graphicx}
\usepackage{textcomp}
\usepackage{xcolor}
\usepackage{algpseudocode}
\usepackage{algorithm}
\usepackage{multirow}
\usepackage{float}
\usepackage{subcaption}
\usepackage{amsthm}
\usepackage{afterpage}
\usepackage{graphicx}
\usepackage{amsmath}
\usepackage{booktabs}
\usepackage{hyperref}
\hypersetup{
    colorlinks=true,
    linkcolor=blue,
    citecolor=blue,
    filecolor=blue,
    urlcolor=black
}
\usepackage[top=0.75in, bottom=1in, left=0.625in, right=0.625in]{geometry}

\algnewcommand\algorithmicforeach{\textbf{for each}}
\algdef{S}[FOR]{ForEach}[1]{\algorithmicforeach\ #1\ \algorithmicdo}
\def\BibTeX{{\rm B\kern-.05em{\sc i\kern-.025em b}\kern-.08em
    T\kern-.1667em\lower.7ex\hbox{E}\kern-.125emX}}
\begin{document}

\title{Synthetic Power Flow Data Generation \color{black} Using \color{black} Physics-Informed Denoising Diffusion Probabilistic Models\\}

	\author{Junfei Wang, 
		Darshana Upadhyay, 
		Marzia Zaman, 
	Pirathayini Srikantha, {\em {Member}},~IEEE
	\thanks{Junfei Wang and Pirathayini Srikantha are with the Department of Electrical Engineering and Computer Science, York University, Toronto, ON, Canada; Darshana Upadhyay is with the Faculty of Computer Science, Dalhousie University, Halifax, Canada; Marzia Zaman is with Cistel Technology, Ottawa, Canada; E-mails: junfeiw@yorku.ca, psrikan@yorku.ca, \color{black}darshana@dal.ca, \color{black} Marzia@cistel.com.}
}

\maketitle

\begin{abstract}
Many data-driven modules in smart grid rely on access to high-quality power flow data; however, real-world data are often limited due to privacy and operational constraints. This paper presents a physics-informed generative framework based on Denoising Diffusion Probabilistic Models \textcolor{black}{(DDPMs)} for synthesizing feasible power flow data. By incorporating auxiliary training and physics-informed loss functions, the proposed method ensures that the generated data exhibit both statistical fidelity and adherence to power system feasibility. We evaluate the approach on the IEEE 14-bus and 30-bus benchmark systems, demonstrating its ability to capture key distributional properties and generalize to out-of-distribution scenarios. Comparative results show that the proposed model outperforms three baseline models in terms of feasibility, diversity, and accuracy of statistical features. This work highlights the potential of integrating generative modelling into data-driven power system applications.
\end{abstract}


\section{Introduction}\label{sec:intro}
The power system is witnessing \color{black}a \color{black} transition from a traditional physical system to a more intelligent and resilient cyber-physical system. This transition is driven by several global challenges, including the environmental impact, the long term increasing trend of electricity demand, the integration of renewable energy resources and the security and privacy concerns, etc. Unlike traditional power systems, which are often centralized and static, smart grid is the modernized power system that incorporates information and communications technology into every aspect of electricity generation, delivery and consumption in order to minimize environmental impact, enhance markets, improve reliability and service, and reduce costs and improve efficiency \cite{IEC}. The availability of high-quality power flow data is essential to many modules in smart grid such as optimal power flow, state estimation, fault detection, EV charging coordination, etc. However, real-world power flow data are often limited, confidential, or subject to privacy constraints, posing challenges to research and innovations. 

To address these issues, the synthetic power flow data generation technique has emerged as a promising approach to support research on the Smart Grid. Traditionally, many studies employ Monte Carlo simulation to generate power flow datasets, where iterative solvers such as the Gauss-Seidel and Newton-Raphson methods are used to solve the power flow equations (e.g., \cite{Wang2022, Liu2018}). However, this approach often involves randomly sampling active power generation and voltage magnitude at all generator buses, which may deviate from real-world data distributions and can render the power flow problem unsolvable in large-scale power grids.

Recently, generative models have shown great potential in synthesizing data without relying on traditional numerical solvers \cite{kingma_vae,gan,Ho2020}. These models learn from real-world datasets and then generate new samples that follow the same underlying distribution as the original data. Moreover, most generative modelling research has focused on sensory data (e.g., images, audio, video) \cite{Ho2020,Kumar2019, Ho2022}, where the generated outputs are considered realistic as long as they are coherent with human perception. However, because physical data must adhere to underlying physical laws and hard constraints, the use of synthetic data in power system research remains an emerging field. In recent years, there has been growing research interest in applying generative models to power systems. The work in \cite{Kababji2020} uses a conditional Generative Adversarial Network (GAN) to generate synthetic smart meter data that captures load patterns and usage habits through a non-intrusive approach. A representation learning method based on an information-theoretically aided GAN is proposed in \cite{Wang2022} to generate candidate Optimal Power Flow (OPF) solutions. \textcolor{black}{A Variational Autoencoder (VAE)} based algorithm is introduced in \cite{Loschenbrand2021} to train a probabilistic deep learning model for solving stochastic OPF under high uncertainty. Paper \cite{Wangc2022} investigates the performance of conditional VAE models in generating a wide range of multivariate load state data. Reference \cite{Guha2023} leverages \textcolor{black}{a} recurrent VAE to detect anomalous voltage data in an unsupervised learning framework. A conditional GAN-based state estimation algorithm is proposed in \cite{He2020}, addressing cases where raw system measurements are incomplete, corrupted, or noisy. A GAN-based security assessment framework is introduced in \cite{Ren2019}, where GANs are trained to fill missing data before making decisions. Paper \cite{Pinceti2021} employs a conditional GAN to produce transmission-level synthetic load data, conditioned on seasonal patterns and node types. While these techniques have shown promise, they often face challenges in accurately modelling the complex, high-dimensional distributions inherent in power systems. For example, GANs are prone to mode collapse, while VAEs may generate blurred or unrealistic samples. Furthermore, important aspects such as adherence to physical laws and constraints, as well as the ability to capture real data distributions, are still open questions in the context of generative modelling for power systems.

Recently, Denoising Diffusion Probabilistic Models (\textcolor{black}{DDPMs}) have gained attention as a powerful alternative for generative tasks. DDPMs leverage a multi-step denoising process to gradually transform a noise distribution into a data distribution, achieving state-of-the-art performance in data synthesis. The key advantage of DDPMs lies in their ability to learn complex data distributions with high resolution. 
\textbf{\textit{Novelty:}} This paper explores the potential of DDPMs to generate physics-informed power flow data. By incorporating physical information into the diffusion process, the proposed model is enabled to capture the intricate relationships between voltage, power generation, and power demands. 
\color{black}
\subsection{Major Contributions}
The following outlines the major contributions of this work:
\begin{itemize}
    \item We develop a physics-informed generative framework using DDPMs tailored for power flow data synthesis.
    \item We incorporate auxiliary training and physics-informed loss functions to ensure physical feasibility and statistical realism in the generated data.
   \item We evaluate our approach on IEEE 14-bus and 30-bus systems and compare our method against baseline DDPMs and physics-informed GAN models across key metrics.
\end{itemize}
\subsection{Outline of the Paper}
\color{black}
In the \textcolor{black}{remainder} of this paper, we first outline the background of power flow study and the fundamental concepts behind DDPMs in Sec.~\ref{sec:form}. Then, we demonstrate how to adapt DDPMs to the power flow domain by incorporating domain-specific constraints in Sec.~\ref{sec:method}. Subsequently, we evaluate the performance of the proposed approach, highlight the advantages and challenges of using DDPMs for synthetic data generation in power systems in Sec.~\ref{sec:ex}. Finally, this work is concluded in Sec.~\ref{sec:conc}.
\vspace{5pt}
\section{Background}\label{sec:form}
This section first introduces the formulation and constraints used in power flow studies, \textcolor{black}{followed by a discussion of the background knowledge on DDPMs.}
\subsection{Power Flow Studies}
Power flow studies are essential for planning, designing, and monitoring power systems, as well as for determining optimal operational strategies. The fundamental information obtained from power flow studies includes active and reactive power generation (denoted as $P_{G_{i}}$ and $Q_{G_{i}}$), active and reactive power demand (denoted as $P_{D_{i}}$ and $Q_{D_{i}}$), and the voltage magnitude $|V_{i}|$ and phase angle $\phi_{i}$ at each bus (node) $i$ in the power grid \cite{Grainger99}. \textcolor{black}{The set of all buses are denoted by $\mathcal{N}$ in this paper, and the set of generator buses by $\mathcal{G}$. The line parameters of the power system are encoded in the nodal admittance matrix $Y$.}

\textcolor{black}{The feasible set in power flow studies comprises all combinations of variables defined above that satisfy the following constraints:}

{\footnotesize
	\begin{align*}
		\text{s.t.} \  \forall  \ i \in \mathcal{G}: \ &P_{G_{i}}^{min}\leq P_{G_{i}} \leq P_{G_{i}}^{max} & \textbf{[C1]} \\
		&Q_{G_{i}}^{min}\leq Q_{G_{i}} \leq Q_{G_{i}}^{max} & \textbf{[C2]} \\
		\text{s.t.} \  \forall  \ i \in \mathcal{N}: &|V_i^{min}|\leq |V_{i}| \leq |V_i^{max}| & \textbf{[C3]}\\
		& \phi_i^{min}\leq \phi_{i} \leq \phi_i^{max}& \textbf{[C4]} \\
		& P_{G_{i}}-P_{D_{i}}+j(Q_{G_{i}}-Q_{D_{i}}) =V_{i}\sum_{j\in \mathcal{N}}V_{j}^{*}Y_{ij}^*    & \textbf{[C5]}\\
		&{|V_{i}(V_{i}^{*}-V_{j}^{*})Y_{ij}^*|\leq S_{i,j}^{max} \ \forall \  j \in \mathcal{N},j\ne i }& \textbf{[C6]}
	\end{align*}
}
\textcolor{black}{in which \textbf{[C1]}–\textbf{[C4]} are inequality constraints of active power generations, reactive power generations, voltage magnitudes and phase angles, respectively.} Furthermore, the physical interactions among these quantities are governed by the power flow equations defined in \textbf{[C5]}, while the maximum allowable power flow through each transmission line is constrained in \textbf{[C6]}. In both \textbf{[C5]} and \textbf{[C6]}.

The general practice in power flow studies is to identify three types of buses in the network, i.e., slack bus, load bus and generator bus. At each bus $i$ two of the four quantities $|V_i|$, $\phi_{i}$ , $P_i=P_{G_{i}}-P_{D_{i}}$ and $Q_i=Q_{G_{i}}-Q_{D_{i}}$ are specified and the remaining two are calculated. The goal of generating synthetic power flow data is to satisfy all physical constraints while ensuring that no consumer or node privacy from the power system is disclosed. As a result, it can be safely released and utilized in both academia and industry.

\subsection{Denoising Diffusion Probabilistic Models}
\textcolor{black}{DDPMs}, originally proposed in \cite{Ho2020}, learn to generate synthetic data by iteratively denoising the data over a series of time steps. They consist of two processes, i.e., the forward diffusion process and the reverse learning process. The basic idea of the forward process is inspired by non-equilibrium thermodynamics, where initial data (a non-equilibrium state) evolves toward an equilibrium state through a stochastic process \cite{Sohl2015}. This process can be mimicked by gradually adding scheduled Gaussian noise (with different mean and variance at each step) to the original data over $T$ time steps until the data loses its meaningful information and becomes noise following standard Normal distribution $N(0,1)$. In the reverse learning process, a neural network learns to reverse this process, transforming noisy data from standard Normal noise back to the original data distribution in $T$ steps.

\subsubsection{Forward Process}
In the forward process of DDPMs, Gaussian noise with scheduled mean and variance is gradually added to the input data along $T$ time steps. This can be defined as in Eq.~\ref{eq:frd}, where $\beta_{t}$ is the schedule parameter at step $t$.
\begin{equation}\label{eq:frd}
	q(x_{t}|x_{t-1})= N(x_{t};\sqrt{1-\beta_{t}}x_{t-1},\beta_{t}I)
\end{equation}
As noted in \cite{Ho2020}, the assumption of Markov Property is made in DDPMs, meaning that the data $x_{t}$ at any step $t$ only depends on the data $x_{t-1}$ at its previous step. By defining a cumulative noise scheduling parameter $\bar{\alpha}_t=\prod_{\tau=1}^{t}(1-\beta_{\tau})$, $x_t$ can be directly sampled once from $x_0$ in Eq.~\ref{eq:frd_2}. 
\begin{equation}\label{eq:frd_2}
		q(x_{t}|x_{0})= N(x_{t};\sqrt{\bar{\alpha}_{t}}x_{0},\bar{\alpha}_{t}I)
\end{equation}
By applying reparamerization trick \cite{kingma_vae}, the expression of $x_t$ can be rewritten in Eq.~\ref{eq:xt}.
\begin{equation}\label{eq:xt}
	x_{t}=\sqrt{\bar{\alpha}_{t}}x_{0}+\sqrt{1-\bar{\alpha}_{t}}\epsilon
\end{equation}
In practice, $\beta_{t}$ is carefully chosen to increase linearly, ensuring $\bar{\alpha}_1\approx 1$ and $\bar{\alpha}_T\approx 0$. This settings guarantee $q(x_1|x_0)$ gradually losing information and $q(x_T|x_0)$ being almost in Normal distribution. E.g., in the original scheduling settings, $\beta_{t}$ linearly increases from $\beta_{1}=1e^{-4}$ to $\beta_{T}=0.2$ over $T=1000$ steps. In this configuration,  $\bar{\alpha}_1=0.9999$ and $\bar{\alpha}_T= 4e^{-5}$. 

\subsubsection{Reverse Learning Process}
Similar to VAE \cite{kingma_vae}, the noise $x_T$ in DDPMs is treated as a latent variable. Theoretically, the objective of DDPMs is to maximize the likelihood of observing each data point in the training dataset under the i.i.d assumption \cite{deep}. However, computing this likelihood involves evaluating an intractable conditional distribution $q(x_{t-1} \mid x_t)$, which depends on the entire training dataset. Thus, another distribution $p(x_{t-1}|x_{t})$ is used to approximate $q(x_{t-1}|x_{t})$. By leveraging the Variational Lower Bound (VLB) \cite{Luo2022}, DDPMs minimize the KL-Divergence between the distribution $p(x_{t-1}|x_{t})$ and $q(x_{t-1}|x_{t}x_{0})$ as defined in Eq.~\ref{eq:goal_gaussian}.
\begin{equation}\label{eq:goal_gaussian}
	\begin{split}
	q(x_{t-1}|x_{t}x_{0})=&N(x_{t-1};\frac{1}{\sqrt{\alpha_{t}}}(x_{t}-\frac{1-\alpha_{t}}{\sqrt{1-\Bar{\alpha}_{t}}}\epsilon),\\
	&\frac{(1-\alpha_{t})(1-\Bar{\alpha}_{t-1})}{1-\Bar{\alpha}_{t}}\mathbf{I})
\end{split}
\end{equation}
In practice, as shown in \cite{Ho2020}, DDPMs work well on a simplified version of the Loss function as defined in Eq.~\ref{eq:simple}, where $\epsilon_\theta(\cdot)$ is a neural network parameterized by $\theta$ to recover the sampled noise $\epsilon_t$. 
\begin{equation}\label{eq:simple}
	\begin{split}
	L_{\text{DDPM}} &= \mathbb{E}_{t \sim [1, T], \mathbf{x}_0, \boldsymbol{\epsilon}_t} \| \boldsymbol{\epsilon}_t - \epsilon_\theta(\mathbf{x}_t, t) \|^2\\
	&= \mathbb{E}_{t \sim [1, T], \mathbf{x}_0, \boldsymbol{\epsilon}_t} \| \boldsymbol{\epsilon}_t - \epsilon_\theta(\sqrt{\bar{\alpha}_t} \mathbf{x}_0 + \sqrt{1 - \bar{\alpha}_t} \boldsymbol{\epsilon}_t, t) \|^2
	\end{split}
\end{equation}
The reverse process iterates from \(t = T\) to \(t = 1\) using the output of \(\epsilon_\theta\) to reconstruct the data:
\begin{equation}
\mathbf{x}_{t-1} = \frac{1}{\sqrt{\alpha_t}} \left(\mathbf{x}_t - \frac{1 - \alpha_t}{\sqrt{1 - \bar{\alpha}_t}} \epsilon_\theta(\mathbf{x}_t, t)\right) + \sigma_t \mathbf{z}
\end{equation}
where \(\mathbf{z} \sim \mathcal{N}(0, I)\) represents Gaussian noise added at each step and \(\sigma_t^2=\beta_{t}\) is the variance that controls the noise level during the reverse process. The model iteratively denoises the noisy data \(\mathbf{x}_t\) and construct less noisy data $x_{t-1}$, gradually moving from standard normal noise at \(t = T\) to a clean data sample at \(t = 0\).

\vspace{5pt}
\section{Proposed Algorithm}\label{sec:method}
This section presents the basic formulation of the physics-informed DDPMs using the original scheduling method, followed by a discussion of its limitations. Subsequently, we propose a data-driven framework to learn scheduling parameters that lead to improved performance in synthesizing power flow data.
\subsection{Physics-informed \textcolor{black}{DDPMs}}
In power flow studies, the key criterion for feasibility is the power balance, defined in \textbf{[C3]}, which includes both active and reactive power balance. A natural approach to incorporate physical knowledge into \textcolor{black}{generative models} is by adding a physics-informed loss to the synthetic data, as in physics-informed GANs \cite{Wang2022}. However, diffusion models are trained in a stochastic manner, based on the Markov chain property. As a result, the physics-informed loss can only be applied at the final step of the reverse process, rather than at all steps.

\begin{figure}[tb]
	\centerline{\includegraphics[scale=0.45]{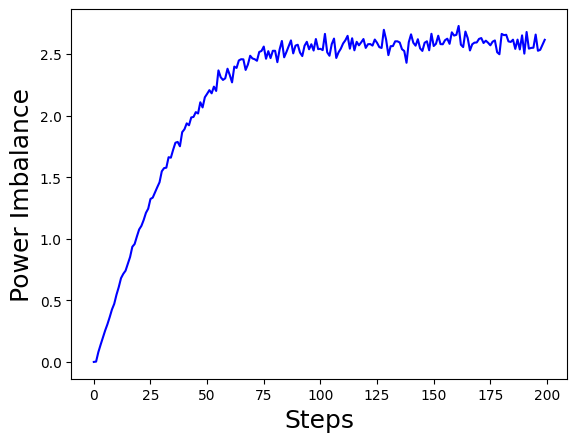}}
	\caption{Average Power Imbalance in Forward Process on IEEE 14-bus System}
	\label{fig:original_imb}
\end{figure}

If $x_{t}=\{P_{G_{i,t}},Q_{G_{i,t}},P_{D_{i,t}},Q_{D_{i,t}},|V_{i,t}|,\phi_{i,t},i\in\mathcal{N}\}$ represents a noisy power flow data at time step $t$, its power imbalance can be defined in Eq.~\ref{eq:residual}, where the average norm of imbalance across all buses are evaluated.
\begin{equation} \label{eq:residual}
	\begin{split}
		R(x_{t})=\mathbf{E}_{i\in \mathcal{N}}|&P_{G_{i,t}}-P_{D_{i,t}}+j(Q_{G_{i,t}}-Q_{D_{i,t}})\\
		&-V_{i,t}\sum_{j\in \mathcal{N}}V_{j,t}^{*}Y_{ij}^*|
	\end{split}
\end{equation}
One of important observations of the forward diffusion process of power flow data is that $R(x_{t})$, which is the violation of physical constraints, is also gradually increasing in the forward process as shown in Fig.~\ref{fig:original_imb}.

Although capturing exact value of power imbalance $R(x_{t})$ through forward diffusion process is impossible, an empirical expectation of imbalance value denoted as $\gamma_{t}$ at time step $t$ on this diffusion curve can be marked. By using $\gamma_{t}$ as an upper bound of imbalance at step $t$, the reverse learning process of DDPMs is guided with the physics-informed knowledge during stochastic training. Accordingly, a physics-informed loss function $L_{R}(x_{t})$ defined in Eq.~\ref{eq:pi_loss} at each step $t$ can be added to the stochastic training process. 
\begin{equation} \label{eq:pi_loss}
	\begin{split}
		L_{R}(x_{t})=\max(R(x_{t})-\gamma_{t},0)
	\end{split}
\end{equation}

The proposed loss function for DDPMs is defined in Eq.~\ref{eq:final_loss}, where $\eta$ is the weight for the physics-informed loss part.
\begin{equation}\label{eq:final_loss}
	L = L_{\text{DDPM}}+\eta L_{R}(x_{t})
\end{equation}
Nevertheless, another important observation is that the loss function $L_{R}(x_{t})$ may not evenly distributes across the all T steps. For example, in Fig.~\ref{fig:original_imb}, the power imbalance under the original scheduling setting of DDPMs rises quickly in the first half of the forward process, but then fluctuates inside a narrow range. From the perspective of power imbalance, the second half of forward diffusion process already lacks meaningful physical structure just like noise. Considering the reverse learning process of DDPMs under this setting, half of the total $T$ steps will not be physically informed, while the remaining steps face the challenging task of significantly reducing the imbalance at each reverse step.

\subsection{Learning Schedule Parameter of \textcolor{black}{DDPMs}}
This section aims to address two key objectives: first, to characterize the desired behaviour of an ideal forward diffusion curve; and second, to explore how such behaviour can be captured through a learnable diffusion parameter framework.

\subsubsection{Linear Diffusion for Power Imbalance}
Assuming the observed power imbalance of Gaussian noise is $\gamma_{T}$, the ideal power imbalance at time step $t$ is defined in Eq.~\ref{eq:rt}, in which the imbalance linearly distribute along the forward diffusion process.
\begin{equation}\label{eq:rt}
	\gamma_{t}=t\frac{\gamma_{T}}{T}
\end{equation}
Despite this complexity, a linear distribution of imbalance is highly desirable for the reverse diffusion process. A gradual and controlled increase in imbalance helps ensure that the reverse process remains both physically informed and stable at each step. In contrast, if the imbalance grows too rapidly or unevenly, the reverse process may struggle to reconstruct realistic power flow states from noisy data. However, due to the highly nonlinear and nonconvex nature of the power flow problem, it is infeasible to manually define numerical scheduling parameters that produce a linear forward diffusion process.

\subsubsection{Learning Scheduling Parameter via Auxiliary Training}\label{sec:aux}
To obtain scheduling parameters that exhibit the desired behavior in the forward diffusion process, we propose a machine learning-based method to learn the scheduling parameter $\bar{\alpha}_{t},\ t \in [0, T]$, such that power imbalance is linearly distributed across all time steps. This approach effectively informs the reverse process of the diffusion model through a physical loss applied at each step, thereby achieving improved performance, as demonstrated in Sec.~\ref{sec:ex}. 

\begin{figure}[tb] 
	\centering
	\includegraphics[scale=0.3]{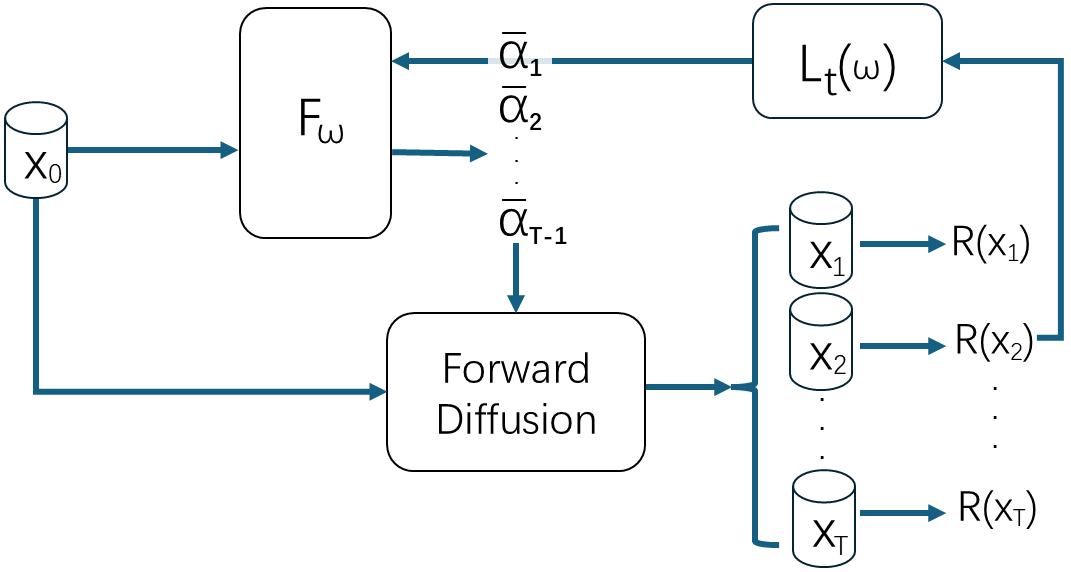}
	\caption{The Framework for Auxiliary Training}
	\label{fig:alphanet}
\end{figure}

The proposed machine learning based framework is illustrated in Fig.~\ref{fig:alphanet}, in which $F_{\omega}$ is a feedforward neural network parameterized by $\omega$. 
The input $X_0$ to $F_{\omega}$ is a set of real power flow data that satisfy all constraints defined in [C1]–[C6], and the outputs are the scheduling parameters $\bar{\alpha}_{t}(\omega)$ at all time steps. These scheduling parameters are then used in the forward diffusion process to generate noisy power flow data at each time step \textcolor{black}{$t\in[1, T]$}, denoted as $X_1, X_2, \ldots, X_T$. Their power imbalance can be evaluated using Eq.~\ref{eq:residual}. The loss function for $F_{\omega}$, defined in Eq.\ref{eq:floss}, measures the mismatch between the expected power imbalance in Eq.\ref{eq:rt} and the actual imbalance $R(X_t)$. This loss function is then used as a penalty in the backpropagation algorithm to update the neural network parameters $\omega$.
\begin{equation}\label{eq:floss}
	L_{t}({\omega}) = |R(\sqrt{\bar{\alpha}_{t}(\omega)}x_{0}+\sqrt{1-\bar{\alpha}_{t}(\omega)}\epsilon)-\gamma_{t}|^{2}
\end{equation}
Unlike traditional neural networks that generate different outputs for different inputs, this model learns a unified scheduling parameter $\bar{\alpha}_{t}(\omega)$ for $t \in [1, T]$ that applies to all data points. This is achieved by computing the mean of the network outputs across the entire real input dataset. This design choice is intentional, as the goal is to ensure consistency in how the diffusion process handles power flow data.

\vspace{5pt}
\section{Experiment}\label{sec:ex}
This section first details the experimental environment used in this work, including the computational platform, software, and benchmark systems. Next, the dataset generation process is described. Subsequently, we evaluate the performance of the proposed framework and compare the results across different configurations of the DDPMs, as well as against another generative model.
\subsection{Environment and Dataset}
In this work, all the experiments ran on Google Colab Cloud Platform with Tesla T4 GPU and 51.0 GB of RAM. Python 3.6 and Tensorflow 2.18.0 are used to implement the algorithms. Two benchmark systems including IEEE 14-bus and IEEE 30-bus systems are used to validate the proposed algorithms. The dataset was generated by MATPOWER's built-in OPF solver\cite{Zimmerman2010}. 

For the IEEE 14-bus system, 70,000 pairs of power flow data were generated for training the DDPMs. Each data entry has \textcolor{black}{60 dimensions}, including 11 each for active and reactive power demand, 14 each for voltage magnitude and phase angle, and 5 each for active and reactive power generation. There are 100K pairs of power flow training data with 112 dimensions for IEEE 30-bus system, containing 20 each for active and reactive power demand, 30 each for voltage magnitude and phase angle, and 6 each for active and reactive power generation. The variability of active and reactive power demands are introduced by randomly perturbing their value in the range of [80\%, 120\%] of their nominal values, which is the common practice in the literature \cite{Liu2022,Pan2021,Wang2024}. Furthermore, the diversity of power flow data is generated by perturbing generator cost coefficient in the range of [50\%, 150\%] of the default values.

\subsection{Performance of the Proposed Model}

\subsubsection{The Performance of Auxiliary Training}
\begin{figure*}[t]
	\centering
	\begin{subfigure}[b]{0.3\textwidth}
		\centering
		\includegraphics[width=\textwidth]{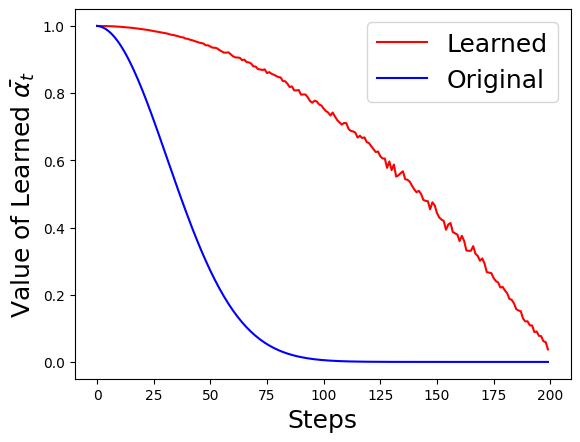}
		\caption{Value of Learned and Original $\bar{\alpha}_t$}
		\label{fig:abnet_alpha_bar}
	\end{subfigure}
	\hfill
	\begin{subfigure}[b]{0.3\textwidth}
		\centering
		\includegraphics[width=\textwidth]{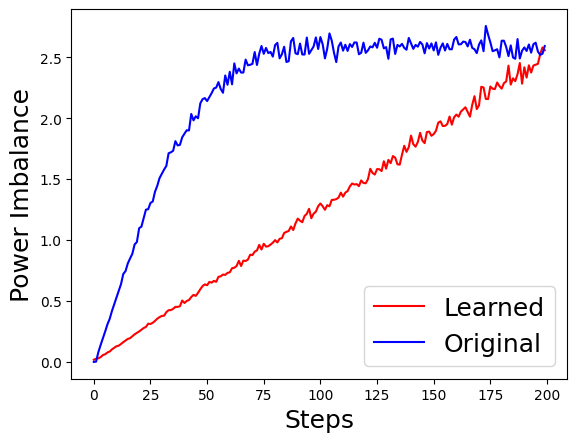}
		\caption{Power Imbalance of Forward Process}
		\label{fig:ab_net_imb}
	\end{subfigure}
	\hfill
	\begin{subfigure}[b]{0.3\textwidth}
		\centering
		\includegraphics[width=\textwidth]{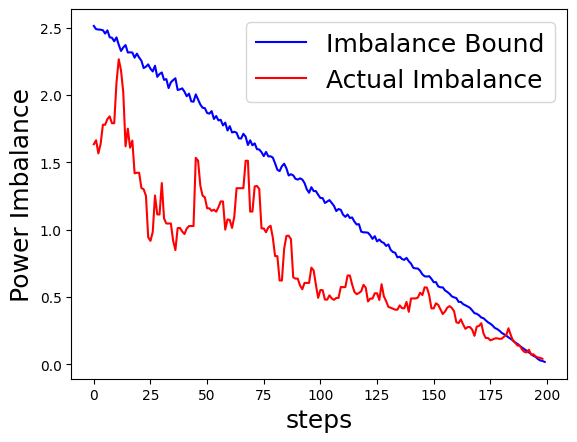}
		\caption{Power Imbalance of Reverse Process}
		\label{fig:ddom_imb}
	\end{subfigure}
	\caption{Performance of Proposed Method on IEEE 14-bus System}
\end{figure*}

For IEEE 14-bus system, the empirical power imbalance of Gaussian noise $\gamma_{T}=2.75$ p.u., while this value is $2.87$ p.u. for IEEE 30-bus system. The total time step $T$ is predefined as 200 for both grids, so the bound for each step $\gamma_{t}$ can be calculated by Eq.~\ref{eq:rt}. As discussed in Sec.~\ref{sec:aux}, the auxiliary training model $F_{\omega}$ generates a set of unified schedule parameters, making it preferable to take the whole dataset as input. However, due to the RAM limitations, batch training is adopted in this paper, and the batch size is selected to be 1,024. The neural networks for both power grids contain two hidden layers with 512 and 256 neurons respectively. The trainable parameters for two neural networks are 217,032 and 240,584 respectively. Fig.~\ref{fig:abnet_alpha_bar} compares the learned and original $\bar{\alpha}_t$ for IEEE 14-bus system. As expected, this learned scheduling parameter linearly distributes imbalance, as shown in Fig.~\ref{fig:ab_net_imb}. 

\subsubsection{The Performance of Physics-Informed DDPMs}
Similar to the original DDPMs \cite{Ho2020} and PixelCNN \cite{Pixelcnn}, a U-Net architecture is selected to train the proposed model. The network consists of seven hidden layers with attention units and time embedding connections. The first three and last three layers serve as down-sampling and up-sampling layers, respectively, while the fourth layer represents the bottleneck. For the two test systems, the number of neurons in the hidden layers is set to [256, 128, 64, 32, 64, 128, 256], resulting in 571,456 and 634,688 trainable parameters, respectively. The output layer uses Sigmoid function to produce values in the range of [0,1], which will be denormalized to feasible region regarding \textbf{[C1]}-\textbf{[C4]}. The parameter $\eta$ in Eq.~\ref{eq:final_loss} is set to 1. \textcolor{black}{For a more comprehensive comparison, we also train physics-informed GANs \cite{Wang2022} on the same datasets, using comparable numbers of parameters.}

A typical reverse process of DDPMs is plotted in Fig.~\ref{fig:ddom_imb}, where the blue curve represents the imbalance bound and the red curve shows the actual power imbalance at each time step. The figure clearly demonstrates that the power imbalance of the synthetic data remains below the imbalance bound throughout most of the $T$ steps, indicating that the reverse process effectively reduces noise and gradually guides the generation toward feasible power flow data. The proposed physics-informed DDPMs produce outputs that not only satisfy all inequality constraints but also exhibit significantly lower power imbalances: 0.013 p.u. for the IEEE 14-bus system and 0.017 p.u. for the IEEE 30-bus system. These residual imbalances can be fully absorbed by slight adjustments to power demand at load buses and generation at generator buses.

\textcolor{black}{We compare the proposed method to three baselines: physics-informed GANs, DDPMs without the physics-informed loss (original setting), and DDPMs with the physics-informed loss but using the original scheduling parameters. The results, presented in Table~\ref{tab:sample}, show that the proposed model and the physics-informed GAN outperform the other two models, both achieving an average power imbalance of approximately 0.01~p.u.}

\begin{table}[htbp]
	\caption{\textcolor{black}{Evaluation of Power Imbalance for Different Generative Models}}
	\centering
	\begin{tabular}{ccccccc}
		\toprule
		\multirow{2}{*}{Model} &GANs&\text{DDPMs}  & \text{DDPMs+Physics} & \text{Proposed} \\
		&+Physics &\text{No Physics} & \text{with original} $\bar{\alpha}_t$ & DDPMs \\
		\midrule
		\multirow{1}{*}{Bus-14} & \textcolor{black}{0.009 p.u.}&0.25 p.u. &  0.11 p.u. & 0.013 p.u. \\
		\multirow{1}{*}{Bus-30} & \textcolor{black}{0.015 p.u.}&0.34 p.u. & 0.09  p.u.&0.017 p.u.\\
		\bottomrule
	\end{tabular}
	\label{tab:sample}
\end{table}

\begin{figure}
	\centering
	\includegraphics[scale=0.65]{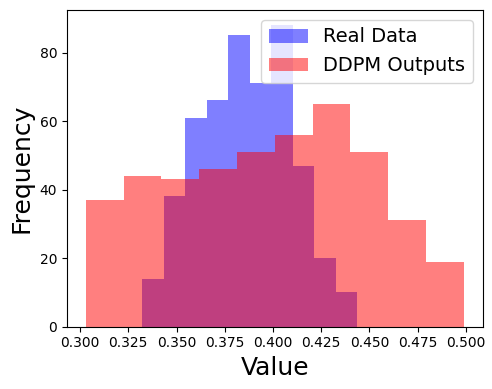}
	\caption{The Diversity of Synthetic Data from Proposed DDPMs}
	\label{fig:hist}
\end{figure}

\begin{figure}
	\centering
	\includegraphics[scale=0.65]{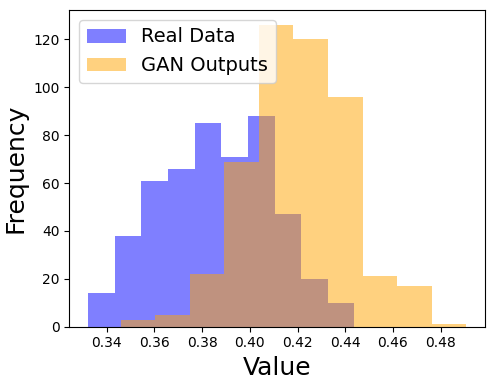}
	\caption{The Diversity of Synthetic Data from GAN}
	\label{fig:gan}
\end{figure}

To evaluate the diversity of the generated data, we compare histograms of synthetic and real samples. \textcolor{black}{As an example, for power generator 3 in the IEEE 14-bus system, shown in Fig.~\ref{fig:hist}, five hundreds of data points generated by DDPMs closely match the peak and spread of the real data distribution. In contrast, the synthetic data produced by physics-informed GAN as shown in Fig.~\ref{fig:gan} do not align well with the real data (shown in blue).} These results highlight the better fidelity of the proposed physics-informed DDPM, both in terms of mean alignment and overall distributional shape.

As a further point of discussion, we observed that the DDPMs are able to learn the non-overlapping regions in Fig.~\ref{fig:hist}, which can be attributed to the incorporation of the physics-informed loss. These regions are also feasible but only exist as long tails in real data set. This observation suggests that the incorporation of physics knowledge encourages the DDPMs to explore beyond the empirical data distribution, enhancing its generalization to out-of-distribution scenarios.

\vspace{5pt}
\section{Conclusion and Future Work}\label{sec:conc}
In this work, a physics-informed generative framework based on DDPMs is proposed for synthesizing feasible power flow data. By incorporating physics-informed loss function through an auxiliary training, the model successfully generates high-fidelity samples that preserve physical feasibility while maintaining diversity. Experiments on the IEEE 14-bus and 30-bus systems show the efficacy of the proposed framework compared to other settings of DDPMs and baseline generative models such as GANs. This framework provides a promising direction for generating secure and privacy-preserving datasets for data-driven power system applications.

Future research in this area can extend this work in several directions. First, since DDPMs generate synthetic data across multiple time steps, improving the speed of DDPMs is an important consideration, especially for real-time or large-scale applications. Second, as power grid size increases, the training of DDPMs may face challenges, exploring distributed or federated learning frameworks with guaranteed feasibility is a promising direction. Third, data generation based on scenario conditions such as weather conditions or time of day could enhance the practical usage of the data for utility companies and time-series studies. Finally, integrating the generative model with optimization modules will be a potential opportunity for both academic research and practical deployment in power system operations.

\end{document}